\documentclass{article}

\usepackage[preprint]{neurips_2024}

\usepackage[utf8]{inputenc}
\usepackage[T1]{fontenc}    
\usepackage{hyperref}       
\usepackage{url}            
\usepackage{booktabs}       
\usepackage{amsfonts}       
\usepackage{nicefrac}       
\usepackage{microtype}      
\usepackage{xcolor}         

\usepackage{amsmath}
\usepackage{amssymb}
\usepackage{mathtools}
\usepackage{amsthm}
\usepackage{pdfpages}

\usepackage[bottom]{footmisc}

\title{GANetic Loss for Generative Adversarial Networks with a Focus on Medical Applications}

\author{
  Shakhnaz~Akhmedova \\
  Centre for Artificial Intelligence in \\ 
  Public Health Research\\
  Robert Koch Institute\\
  Berlin, Germany\\
  \texttt{AkhmedovaS@rki.de} \\
   \And
   Nils~K\"orber  \\
  Centre for Artificial Intelligence in \\
  Public Health Research \\
  Robert Koch Institute\\
  Berlin, Germany\\
  \texttt{KoerberN@rki.de} \\
}

\begin{document}

\maketitle

\begin{abstract}
\newcommand\blfootnote[1]{
  \begin{NoHyper}
  \renewcommand\thefootnote{}\footnote{#1}
  \addtocounter{footnote}{-1}
  \end{NoHyper}
}

Generative adversarial networks (GANs) are machine learning models that are used to estimate the underlying statistical structure of a given dataset and as a result can be used for a variety of tasks such as image generation or anomaly detection. Despite their initial simplicity, designing an effective loss function for training GANs remains challenging, and various loss functions have been proposed aiming to improve the performance and stability of the generative models. In this study, loss function design for GANs is presented as an optimization problem solved using the genetic programming (GP) approach. Initial experiments were carried out using small Deep Convolutional GAN (DCGAN) model and the MNIST dataset, in order to search experimentally for an improved loss function. The functions found were evaluated on CIFAR10, with the best function, named GANetic loss, showing exceptionally better performance and stability compared to the losses commonly used for GAN training. To further evalute its general applicability on more challenging problems, GANetic loss was applied for two medical applications: image generation and anomaly detection. Experiments were performed with histopathological, gastrointestinal or glaucoma images to evaluate the GANetic loss in medical image generation, resulting in improved image quality compared to the baseline models. The GANetic Loss used for polyp and glaucoma images showed a strong improvement in the detection of anomalies. In summary, the GANetic loss function was evaluated on multiple datasets and applications where it consistently outperforms alternative loss functions. Moreover, GANetic loss leads to stable training and reproducible results, a known weak spot of GANs.\blfootnote{Code: \url{https://github.com/ZKI-PH-ImageAnalysis/GANetic-Loss}.}

\end{abstract}

\section{Introduction}

\begin{figure}
  \centering
  \includegraphics[width=0.85\linewidth]{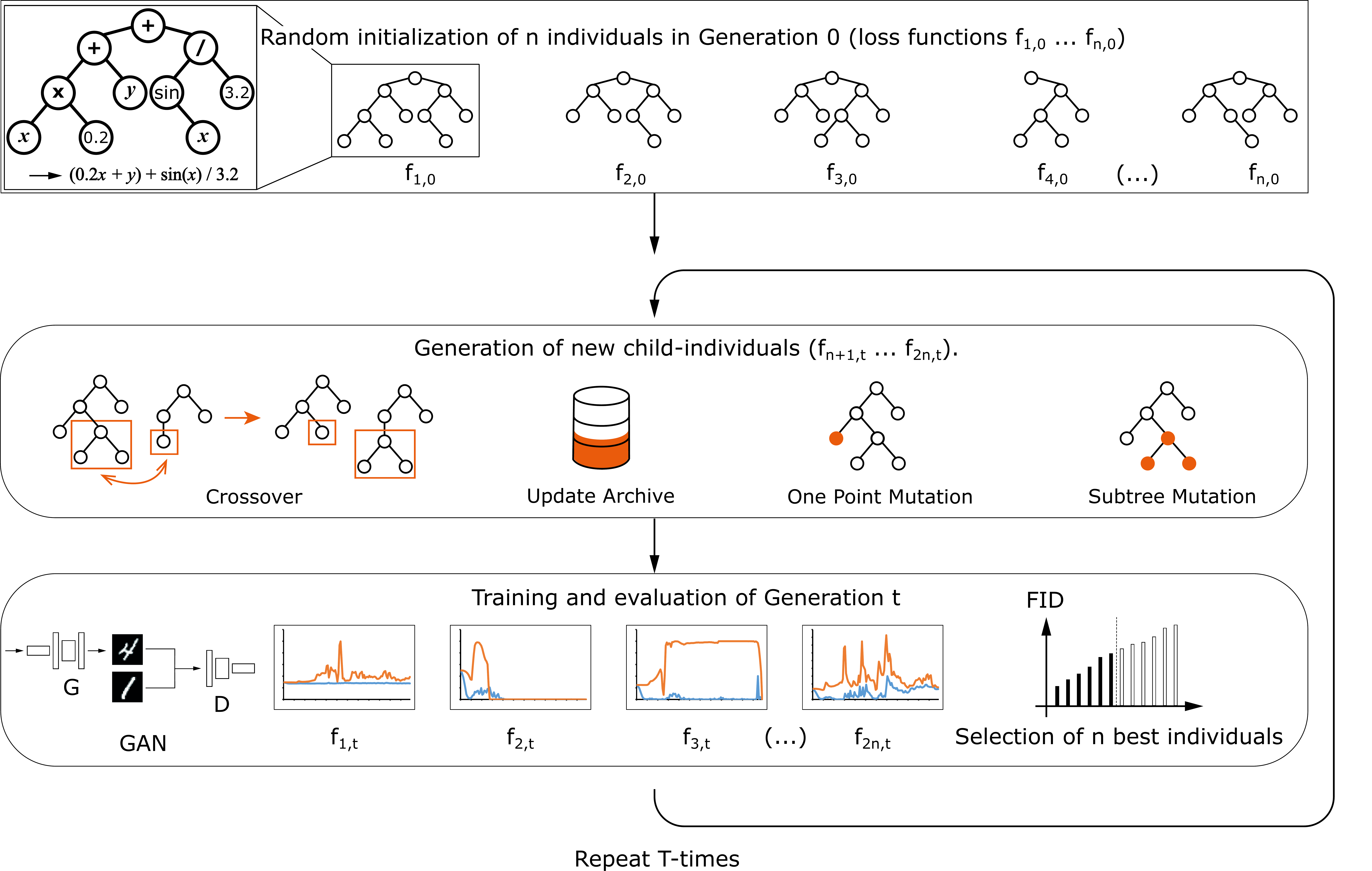}
  \caption{Implemented GP search process. In the first row an example of the generated trees is shown, in this example $x$ and $y$ are $y_{real}$ and $y_{pred}$.}
\label{fig:fig_1}
\vskip -0.25in
\end{figure}

Generative Adversarial Networks (GANs) were first introduced in~\cite{Goodfellow14} as a new framework for estimating generative models via an adversarial process. Specifically, the GAN model is designed as two competing neural networks that are trained simultaneously, with the first network referred to as generator $G$ and the second as discriminator $D$. The goal of the generator is to capture the data distribution in such a way that it is able to create data patterns that are as similar as possible to those in the training data. On the other hand, the discriminator tries to estimate the probability that a sample comes from the training data and was not generated by $G$. 

GANs have been utilized in the generation of images~\cite{Bousmalis16, Ledig16}, video~\cite{Tulyakov17, Walker17}, audio~\cite{Lee17, Guimaraes17} and text~\cite{Guo17, Autume19}. Moreover, GANs can be applied for different problems in the machine learning field such as anomaly detection~\cite{Xia22}, semi-supervised learning~\cite{Dai17, Gao21}, representation learning~\cite{Yuan20, Tran17}, which attempts to address the lack of labeled data and learn representations of the data to facilitate the extraction of useful information related to the task.

Originally it was proposed to train GANs by using the adversarial loss function~\cite{Goodfellow14}. However, several studies have shown that these GAN models are prone to mode collapse, show unstable results as they fail to converge in some cases, and consequently do not produce realistic images~\cite{Che16, Richardson18, Metz16}. 

There are various variants of GAN models, which differ from each other either by the layers, overall structure of the generator and/or discriminator networks, or by the incorporated loss function used for training. However, the training process of GANs is very complex and often unstable, so an effective loss function can significantly improve the efficiency of GAN models. Thus, the focus of this study is to find a loss function that is generally good regardless of the specific GAN architecture and dataset. 

The loss function search can be represented as an optimization problem, that can be solved by using Genetic Programming (GP), which is a well-known population-based evolutionary algorithm~\cite{Augusto00}. The GP approach is frequently used to design previously unknown heuristics in order to solve complex computational search problems~\cite{Burke09}. GP was applied to determine the loss function for a relatively simple Deep Convolutional GAN model (DCGAN)~\cite{Radford16} using MNIST dataset~\cite{lecun1998mnist}. The GP experiments resulted in several loss function candidates potentially suitable for GAN training. 

All loss functions found were used to train a slightly larger DCGAN model with the CIFAR10 dataset and were compared to commonly used loss functions. The best function found, GANetic loss, was used for medical applications, namely image generation and anomaly detection, using state-of-the-art models. A comparison with other commonly used loss functions for GANs showed the efficiency of the GANetic loss, as it was able to outperform known representative baselines in performance and stability.

Thus, the main contributions in this paper can be summarized as follows:
\begin{itemize}
\item{It is proposed to use the GP approach to design the loss function for GANs. The loss function found, GANetic loss, showed more stable training and better results compared to commonly used loss functions.}
\item{The GANetic loss was evaluated by using two different GAN models for medical imaging tasks such as image generation and anomaly detection. Experimental results show that GANetic loss is superior to other loss functions resulting in better image generation and anomaly detection. }
\item{Furthermore, the comparison with different known losses commonly used for training GANs is shown to prove that the new loss function is able to avoid mode collapses, leading to reproducible results.}
\item{Self-regularization naturally implemented in the GANetic loss indicates a direction for loss function design in general.}
\end{itemize}

\section{Related Work}

Generative adversarial networks have gained interest and popularity among researchers due to the relative simplicity of the idea of two competing networks, where one tries to trick another, and their capability to represent complex and high-dimensional data. The GAN architecture, originally proposed in~\cite{Goodfellow14}, consists of two neural networks, generator $G$ and discriminator $D$, competing adversarially against each other. The architecture of these networks can have fully-connected~\cite{Barua19}, convolutional~\cite{Radford16} or recurrent layers~\cite{Esteban17} depending on the data domain and application. The generator $G$ is trained to produce samples which capture the underlying distribution of the training data, while the discriminator $D$ is fed a sample of mixed data in order to estimate the probability of whether the sample is real, i.e. originates from the training dataset, or fake, i.e. was generated by $G$. Thus, GANs minimize the Jensen-Shannon divergence between a target data distribution and the model distribution given by the generator~\cite{Goodfellow14}:

\begin{equation}
\label{eqn:eq1}
\min_{G}\max_{D}V(D,G)=\mathbb{E}_{x \sim P_{data}(x)}[log(D(x)] + \mathbb{E}_{z \sim P_{z}(z)}[1-log(D(G(z))].
\end{equation}

The following notations are used in Equation \ref{eqn:eq1}: $G(z)$ is the data generated by the generator model, $P_{data}(x)$ is the probability distribution of the real samples presented in the training dataset, $P_{z}(z)$ is the probability distribution of input noise variables, $D(x)$ and $D(G(z))$ are the probabilities that $x$ and $G(z)$ are from the training dataset. It should be noted that one of the most commonly used implementations of GANs consists in using the binary cross entropy loss function for discriminator $D$ to update both generator and discriminator after each training step. However, in~\cite{Mao17} authors showed that the sigmoid cross entropy loss function leads to the problem of vanishing gradients, which can cause the model to fail to converge and give poor results. Therefore, it is crucial to properly design the loss function for a chosen GAN model to ensure its stability and efficiency.

In recent years various GAN models have been proposed, which differ from each other either by the architecture of the implemented neural networks~\cite{Liu21, Sauer21}, by applying augmentations~\cite{Karras20} or by the designed loss functions used during the training~\cite{Arjovsky17, Mao17}. The most well-known studies among them, related to the GAN loss function design, are about the Wasserstein loss, that was introduced~\cite{Arjovsky17} for WGANs, and the least squares loss function, which was implemented for LSGANs~\cite{Mao17}. Another alternative, commonly used to train GAN models, is the hinge loss function~\cite{Mustafa20}. However, the choice of loss function usually depends on the experience of the researchers, the model, and the specific problem.

\section{Genetic Programming for Loss Function Design}

Genetic Programming (GP) is a population-based evolutionary algorithm, which can be applied to design different heuristics, depending on the problem at hand, and, additionally, it is considered as a machine learning tool as it can be used to discover a functional relationship between features in data. GP was initially inspired by the biological evolution, including natural processes such as mutation and selection. In GP, each solution is represented by a tree structure, and each of these trees are evaluated recursively to produce the resulting multivariate expression. There are two types of nodes used for tree-based GP: the terminal node, also called leaf, which is randomly chosen from the set of variables, and the tree node, which can be chosen from a predefined set of operators. 

The GP search process starts with random initialization of a set of potential solutions, which is also called a population of individuals, in the functional search space. The number of individuals is the predefined parameter $n$, which does not change through the whole search process. The overall GP search process used here is summarized in Figure \ref{fig:fig_1}. Thus, for this study a set of loss functions was randomly generated, and each loss function was represented as a tree, where the terminals were randomly chosen from the set $\{y_{pred}, y_{real}, \mathbb{R}\}$, while the operators were chosen from the set $\{+, -, \times, \div, \sqrt, \log, \exp, \sin, \cos\}$. To utilize these functions and terminals for loss function generation several modifications, described in Appendix \ref{app_a}, were applied.

After initialization, the main search loop starts, which consists of the iteratively repeated steps crossover, mutation, fitness function evaluation and selection (the number of GP steps or generations is denoted as $T$). The fitness function $F$ determines the quality of the individual and is the crucial part of the search process, as our goal was to experimentally find a robust loss function that can be generalized to a variety of datasets. Specifically, each individual loss function was used to train both generator and discriminator of a simple DCGAN model $5$ times on MNIST dataset~\cite{lecun1998mnist}, and the mean FID value and standard deviation calculated over the $5$ program runs was used as the fitness value.

The crossover operator is used to exchange the subtrees between two individuals (Figure \ref{fig:fig_1}). Mutation can be applied in various ways, but in this study two variants were used: a random subtree in the tree is chosen and replaced with another randomly generated subtree; a random node in the tree is chosen and replaced with another randomly generated node.

Both mutation operators as well as crossover have their own parameters, such as mutation $M_{ST}$, $M_N$ and crossover $Cr$ rates, which determine how often individuals will be changed during the main search loop. Finally, all individuals and the ones generated after crossover and mutation steps are combined into one population, and then the selection operator is applied. Here, two selection schemes were considered: the tournament selection with $k_t$ as tournament size and selection of the most fit individuals from the combined population. 

Moreover, the success history-based adaptation strategy was used to improve the efficiency of the GP approach. To be more specific, at the beginning of the main search loop an empty external archive $A$ was created. The maximum size of this archive was set equal to the population size, $n_A = n$. All individuals not passing selection could have been saved in the archive with a given probability $p_A$. Individuals saved in the external set were used during crossover step with some probability $Cr_A$.

In summary, loss function search was performed by the GP algorithm and essentially can be described as an optimization process. A set of individuals, where each individual is a mathematical formula representing a loss function, is generated. These individuals are modified by crossover and mutation, and the most fit individuals are selected for the next generation. These actions are repeated a certain number of times in order to find the best loss function.

\section{Experiments}

\subsection{Genetic Programming}

As mentioned, a small GAN model was trained on the MNIST dataset for each individual during GP. Each GAN (its generator and discriminator) was trained 5 times with its corresponding generated loss function to calculate the mean and deviation of the FID value. The discriminator of this GAN model consists of two convolutional layers with LeakyReLU activations and no pooling layers, the last convolutional layer was flattened and connected with a dropout~\cite{Srivastava14} of $0.4$ to a dense layer comprising a single output node used to discriminate between real and fake images. 

The architecture of the generator was slightly more complex. A sample of random numbers from the standard normal distribution were given as input to the generator. These values were transformed by reshaping a fully-connected layer, followed by a convolutional layer with LeakyReLU activation into matrices of the size of the training images. Finally, the generated 2D matrices were used as input to an autoencoder whose encoder had the same layers as the discriminator but without flattening or dropout, and a decoder built accordingly in reverse order (a transposed convolutional layer for each convolutional layer of the encoder with a final hyperbolic tangent output activation). The included autoencoder was chosen since most anomaly detection models use similar architecture which was one of the tasks considered in the study. 

Genetic programming algorithm has several parameters, which can affect speed and quality of the search process. The following parameters were the same for all GP experiments: population size $n=10$, number of generations $T=50$, crossover rate $Cr=0.7$, the minimum tree height was set to $2$, while the maximum tree size was $100$. To create leaf nodes real valued terminals were sampled from a uniform distribution in the range $[-5,5]$. In total, we tested $8$ different configurations of the GP algorithm (see Table \ref{tab:tab_1} for details), resulting in $8$ individual experiments, each yielding a sub-optimal solution for the search process. The equations for found loss functions are shown in Equations \ref{eqn:eq2}-\ref{eqn:eq9} (Appendix \ref{app_c}), where $N=2$ is the number of classes.

In order to limit the total GAN training time of the GP search, a batch size of $512$ and $70$ epochs of training were chosen. Each GAN was trained using the Adam optimizer~\cite{Diederik15} for both networks with learning rate and $\beta_1$ of $0.0002$ and $0.5$, respectively, and the latent variable dimension of $100$. To further reduce the evaluation, the FID value was calculated by generating $256$ images from the trained model and randomly selecting the same number from the data set. Each GP experiment took about $100$ hours to complete on an internal server node equipped with $8$ Nvidia A100. The results obtained for the MNIST data set using the best loss functions found in the experiments compared to the binary cross entropy (BCE) function are shown in Table \ref{tab:tab_2}. 

\begin{table}[t]
\caption{Numerical results from MNIST dataset. Mean FID values, standard deviation (STD) and mean discriminator accuracy ($D\textsubscript{Acc}$)  over 5 runs.}
\label{tab:tab_2}
\centering
\begin{tabular}{lccccccccc}
\toprule
Loss & BCE & $f_1$ & $f_2$ & $f_3$ & $f_4$ & $f_5$ & $f_6$ & $f_7$ & $f_8$ \\
\midrule
Mean & 11.59 & 9.86 & 15.00 & 6.54 & \textbf{5.96} & 7.17 & 6.18 & 7.44 & 9.70 \\
STD & 7.77 & 2.29 & 8.55 & \textbf{1.25} & 1.86 & 3.68 & 2.56 & 2.90 & 1.43 \\
$D\textsubscript{Acc}$ & \textbf{98.1} & 2.8 & 4.8 & 17.4 & 94.1 & 49.9 & 96.4 & 49.8 & 46.6 \\
\bottomrule
\end{tabular}
\end{table}

To evaluate the functions found by GP, not only the quality of the generated images was taken into account, but also the accuracy of the discriminator and the variation of individual training runs (explanation for that choice is given in Appendix \ref{app_mnist}). Quality of generated images, measured by FID, compared to the BCE loss, could be improved by all but one found functions. To find a function resulting in stable training, we used the standard deviation for evaluation during GP to take reproducibility into account. As a result the functions found show, except for one, better stability compared to BCE loss. 

From the $8$ functions resulting from GP, $f_4$ and $f_6$ showed the best results, both found using the success history-based adaptation strategy. However, the function $f_4$ showed better results than $f_6$ in terms of FID value and standard deviation. Consequently, only $f_4$, denoted as GANetic loss, was further evaluated and compared to popular loss functions commonly used to train GANs. GANetic loss is defined by the following formula:

\begin{equation}
\label{eqn:GANetic}
GANetic=\frac{1}{N}\sum_{i=1}^{N}{\left(\left(y_{pred}^{(i)}\right)^3+ \sqrt{\left|\alpha \times \frac{y_{real}^{(i)}}{y_{pred}^{(i)}+\varepsilon}\right|+\varepsilon}\right)},
\end{equation}
where $\alpha=3.985$.

\subsection{Evaluation of the GANetic Loss}
\label{cifar10}

The first set of experiments was conducted to evaluate the performance of the GANetic loss function and to compare it to other well-known loss functions commonly used to train GANs. For these experiments we used a slightly bigger GAN and trained it on the CIFAR10 dataset~\cite{krizhevsky2009learning}. Specifically, the model from the previous section was extended to include two additional convolutional layers for the discriminator and the encoder of the generator, and consequently two additional transposed convolutional layers for the decoder.

Each GAN was trained with the batch size of $128$ for $200$ epochs. The results of $25$ independent program runs were compared to the identical model using binary cross entropy, hinge, adversarial, Wasserstein, and least square loss functions. Corresponding FID values (the best, the worst, the mean) and their standard deviation are presented in Table \ref{tab:tab_3}. Examples of the generated images are demonstrated in Figure \ref{fig:fig_cifar10} (Appendix \ref{app_cifar10})

\begin{table}[t]
\caption{FID values on CIFAR10 dataset.}
\label{tab:tab_3}
\centering
\begin{tabular}{lcccccc}
\toprule
Loss & Adversarial & BCE & Hinge & Wasserstein & LeastSquare & GANetic \\
\midrule
Best & 9.45 & 2.06 & 7.11 & 3.04 & 4.60 & \textbf{1.52} \\
Worst & 72.58 & 45.30 & 62.47 & 31.74 & 80.87 & \textbf{2.76} \\
Mean & 18.52 & 8.34 & 12.41 & 12.93 & 17.52 & \textbf{2.20} \\
STD & 16.12 & 8.92 & 10.59 & 4.85 & 15.81 & \textbf{0.39} \\
\bottomrule
\end{tabular}
\end{table}

\begin{figure}[t]

\centering
\includegraphics[width=0.6\linewidth]{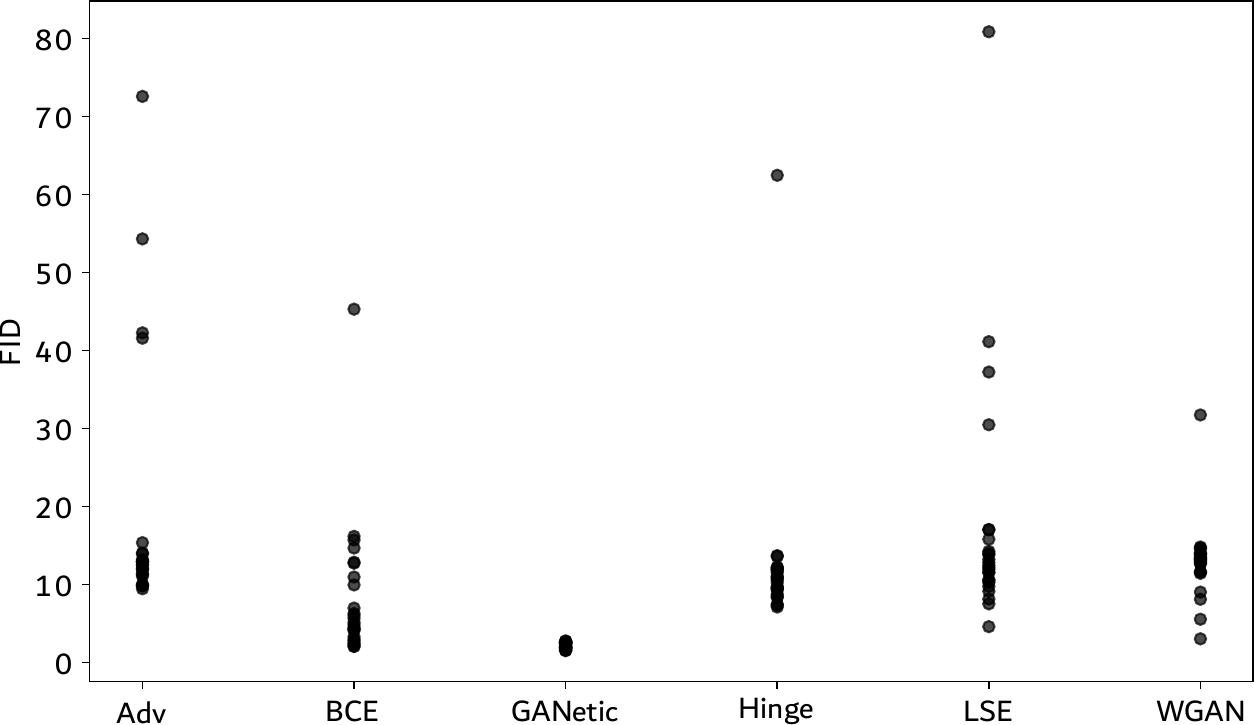}
\caption{Variation of results on CIFAR10 for different losses over 25 runs.}
\label{fig:fig_5}
\end{figure}

From the tested functions GANetic loss showed the most consistent results outperforming all other functions by a large margin. Figure \ref{fig:fig_5} shows the variation of FID values for the tested loss functions over $25$ runs. GANetic loss leads to very reproducible results avoiding mode collapses, which can be observed for all other functions during the $25$ trials.

Finally, two additional configurations of the GANetic loss were tested in an ablation study. For that purpose the same DCGAN model and CIFAR10 dataset were used. We tested configurations in which the generator, the discriminator or both replaced the BCE function with GANetic loss during training. Obtained results are demonstrated in Table \ref{tab:tab_5}.

While most of the performance gains were achieved by using GANetic loss for the discriminator, the overall best performance was achieved when the generator and discriminator were trained with GANetic loss. Results from Table \ref{tab:tab_5} show that DCGAN with both networks, generator and discriminator, trained by using GANetic loss achieves better results in terms of the FID metric value. Unless stated otherwise, all other experiments were conducted using GANetic loss for generator and discriminator.

\subsection{Medical Applications}

\subsubsection{Image Generation}

To test the performance of GANetic loss in a more complex scenario, we used the state-of-the-art projected GAN (prGAN)~\cite{Sauer21} and StyleGAN2-ADA~\cite{Karras20} architectures and three medical datasets, namely BrecaHAD~\cite{Aksac19}, Hyper-Kvasir~\cite{Borgli2020} and LAG~\cite{Li_2019_CVPR}. The GANetic loss was applied to train the generator and discriminator networks implemented for projected GANs (prGANs) and only generator network for StyleGAN2-ADA model. 

The BreCaHAD dataset consist of only 162 breast cancer histopathology images with an image size of $1360 \times 1024$ that were reorganized into 1944 partially overlapping crops of size $512 \times 512$, as implemented in~\cite{Karras20}. The Hyper-Kvasir dataset contains 110,079 gastrointestinal images and 374 videos where it captures anatomical landmarks and pathological and normal findings. Following~\cite{Tian2021ConstrainedCD}, we take 2100 images from \textit{cecum}, \textit{ileum}, \textit{bbps-2–3} and 1000 images from \textit{polyp} using the same subset as used for anomly detection as shown later. The images were combined and reorganized into 10,417 partially overlapping crops, that were resized to $256  \times 256$. The LAG dataset used for glaucoma detection, contains 4854 images with 1711 positive glaucoma and 3143 negative glaucoma scans. For image generation experiments, the images from LAG datasets were center cropped and resized to $256 \times 256$. 

The evaluation protocol of~\cite{Sauer21} (including FastGAN generator) was followed to allow fair comparisons for prGANs. Therefore, the prGANs trained with all the loss functions were given the same fixed number of frames on all datasets, namely $10^7$. It was found that for the BreCaHAD dataset regardless of the loss function, the prGAN model began to overfit after $3\times10^6$ or $4\times10^6$ images and consequently showed poor results, so experiments were repeated with the same parameters but given only $4\times10^6$ images. For the Hyper-Kvasir and LAG datasets experiments were carried out with $10^7$ images. For StyleGAN2-ADA the number of frames was set to $10\times10^7$ for all datasets.

\begin{figure}[t]

\centering
\includegraphics[width=\columnwidth]{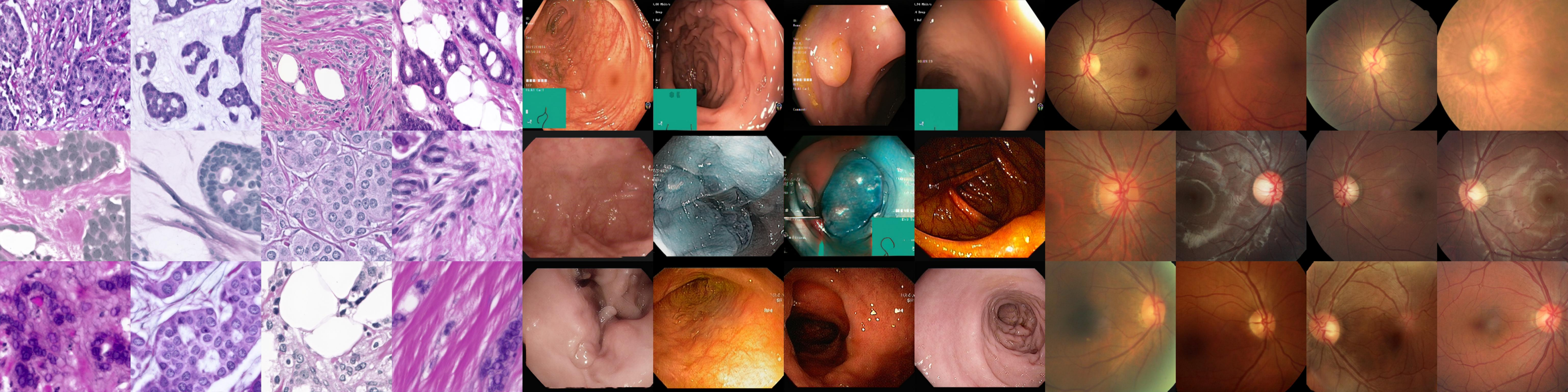}
\caption{Examples of images generated by prGANs using different losses for BreCaHAD (left), Hyper-Kvasir (middle) and LAG (right) datasets. From top to bottom: ground truth images, original loss and GANetic loss.}
\label{fig:fig_6}
\end{figure}

\begin{table}[t]
\caption{FID and KID values for different losses trained on BreCaHAD, Hyper-Kvasir and LAG datasets.}
\label{tab:tab_6}
\centering
\begin{tabular}{lcccccc}
\toprule
 & \multicolumn{2}{c}{BreCaHAD} & \multicolumn{2}{c}{Kvasir} & \multicolumn{2}{c}{LAG} \\
\cmidrule(r){2-7}
Model (Loss) & FID & KID & FID & KID & FID & KID \\
\midrule
prGAN (original) & 15.50 & 3.49 & 4.01 & 1.77 & 3.73 & 0.93 \\
prGAN (GANetic) & \textbf{14.57} & 3.38 & \textbf{3.81} & \textbf{1.16}& \textbf{3.53} & \textbf{0.90}\\
StyleGAN2-ADA (original) & 15.77 & 2.88 & 5.55 & 2.19 & 6.38 & 3.00 \\
StyleGAN2-ADA (GANetic) & 16.2 & \textbf{2.85} & 5.80 & 2.07 & 6.09 & 2.79 \\
\bottomrule
\end{tabular}
\end{table}

The obtained results, FID and KID values between $50k$ generated and all real images, are presented in Table \ref{tab:tab_6}. The results show that GANetic loss leads to an improved performance for prGANs on all datasets and for StyleGAN2-ADA improves at least one metric. In addition, it should be noted that prGANs trained with GANetic loss showed the fastest convergence after approximately $1.8\times10^6$ iterations, while the original configuration required $3.4\times10^6$, and $2.8\times10^6$, respectively, which is demonstrated in Figure \ref{fig:fig_conv} (see Appendix \ref{app_d}). The convergence was similar on the Hyper-Kvasir and LAG datasets regardless of the loss function. Examples of the generated images for both models and all the mentioned datasets are demonstrated in Figures \ref{fig:fig_6} and \ref{fig:fig_style} (Appendix \ref{app_med_image_gen}). Additional evaluation using the same setup on the AFHQ-Cats,AFHQ-Dogs and AFHQ-Wild datasets ~\cite{choi2020stargan} can be found in Appendix \ref{tab:tab_afhq}.

\subsubsection{Anomaly Detection}

The detection of imaging biomarkers correlating with disease status is important for initial diagnosis, assessment of treatment response and follow-up examinations. However, training deep learning models for the identification of imaging biomarkers is diffult as it requires expert annotated data. Even with annotated training data available, supervised learning is limited to already known markers. Thus, in~\cite{Schlegl2019fAnoGANFU} authors proposed a GAN based unsupervised learning approach, called f-AnoGAN, to identify anomalous images and image segments, that can serve as imaging biomarker candidates.

In~\cite{Schlegl2019fAnoGANFU} authors described an anomaly detection framework, which consists of two training steps on normal ''healthy'' images: GAN training and encoder training, where the latter is based on the trained GAN model. After training, inference yields an anomaly score for a new image utilizing these trained components. 

Experiments were carried out on two datasets: Hyper-Kvasir and LAG~\cite{Li_2019_CVPR}. Following~\cite{Tian2021ConstrainedCD}, we took 2100 images from \textit{cecum}, \textit{ileum} and \textit{bbps-2–3} cases as normal and 1000 abnormal images from \textit{polyp} as abnormal. We used 1600 images for the training set and 500 images for the test set similarly to what was done in~\cite{Rafiee2023AbnormalityDF}. For the LAG dataset 2343 normal (negative glaucoma) images were used for training and 800 normal images and 1711 abnormal (positive glaucoma) images were used for testing the trained model. 

All images were resized to $64 \times 64$, and augmentation included horizontal flip and normalization. The f-AnoGAN approach uses an improved WGAN architecture, namely, it is based on the WGAN model with gradient penalty (WGAN-GP)~\cite{Gulrajani2017ImprovedTO}. In this study, the number of epochs for both WGAN-GP and encoder was set to 1000. 

\begin{table}[t]
\caption{AUROC results for Hyper-Kvasir and LAG datasets.}
\label{tab:tab_8}
\centering
\begin{tabular}{lcc}
\toprule
Model & Kvasir & LAG \\
\midrule
DAE~\cite{Masci2011StackedCA} &0.705 & 0.651\\
OCGAN~\cite{Perera2019OCGANON} & 0.813 & 0.737\\
f-anoGAN~\cite{Schlegl2019fAnoGANFU} & 0.907 & 0.778\\
ADGAN~\cite{Liu2019PhotoshoppingCV} & 0.913 & 0.752\\
CAVGA-Ru~\cite{Venkataramanan2019AttentionGA}& 0.928 & 0.819\\
IGD~\cite{Chen2021UnsupervisedAD} & 0.937 & 0.857\\
CCD~\cite{Tian2021ConstrainedCD} & 0.972& \textbf{0.874}\\
\midrule
f-anoGAN (GANetic) & \textbf{0.987} & 0.782\\
\bottomrule
\end{tabular}
\end{table}

Experimental results for both datasets of the f-AnoGAN model, which generator and discriminator (but not encoder) were trained using GANetic loss, are demonstrated in Table \ref{tab:tab_8}. Overall, applying GANetic loss to f-AnoGAN improved model's performance for both datasets. While it only leads to a mediocre improvement for the LAG dataset, it results in a massive performance boost when applied to Hyper-Kvasir dataset outperforming all other methods. 

Results for the remaining configurations of the trained f-AnoGAN model are shown in Table \ref{tab:tab_9} (Appendix \ref{app_e}). Additionally, the discrete distribution of anomaly scores and the AUROC curve for Hyper-Kvasir datasets are shown in Figure \ref{fig:fig_kvasir}  (Appendix \ref{app_e})).

\section{Discussion}

The training of GANs can be viewed as a binary classification problem, where the generator is optimized by making the discriminator predict true labels and the discriminator distinguishes between true and false images. Thus, the GANetic loss function in equation \ref{eqn:GANetic} can be simplified as:

\begin{equation}
\label{eqn:eq5}
GANetic=\left(y_{pred}^{(0)}\right)^3+\left(y_{pred}^{(1)}\right)^3+ \sqrt{\left| \frac{\alpha}{y_{pred}^{(1)}+\varepsilon}\right|+\varepsilon},
\end{equation}

where $\alpha=3.985$, $\varepsilon=10^{-8}$ and $y_{pred}^{\left(i\right)}$, $i \in \{0,1\}$, denotes the model output for $y_{real} \in \{0,1\}$. 

It should be noted that $y_{pred}$ was in the range $[0,1]$, being passed through a sigmoid function. Several conclusions can be drawn from the experimental derived equation. First, predictions for $y_{real} = 0$ are treated differently compared to $y_{real} = 1$. The former showing a cubic, monotonically increasing function, while the latter exhibits a minimum near $0.75$. Morevover, discriminator's penalty for misclassifying the fake images is limited by $1$, while it is much larger for misclassifying real images. This could indicate that it is advantageous to focus on the real images during training and could explain the faster convergence of GANetic loss, as the generator normally requires a considerable number of images to produce reasonably realistic images. 

\begin{figure}[t]
\centering
\includegraphics[width=0.9\columnwidth]{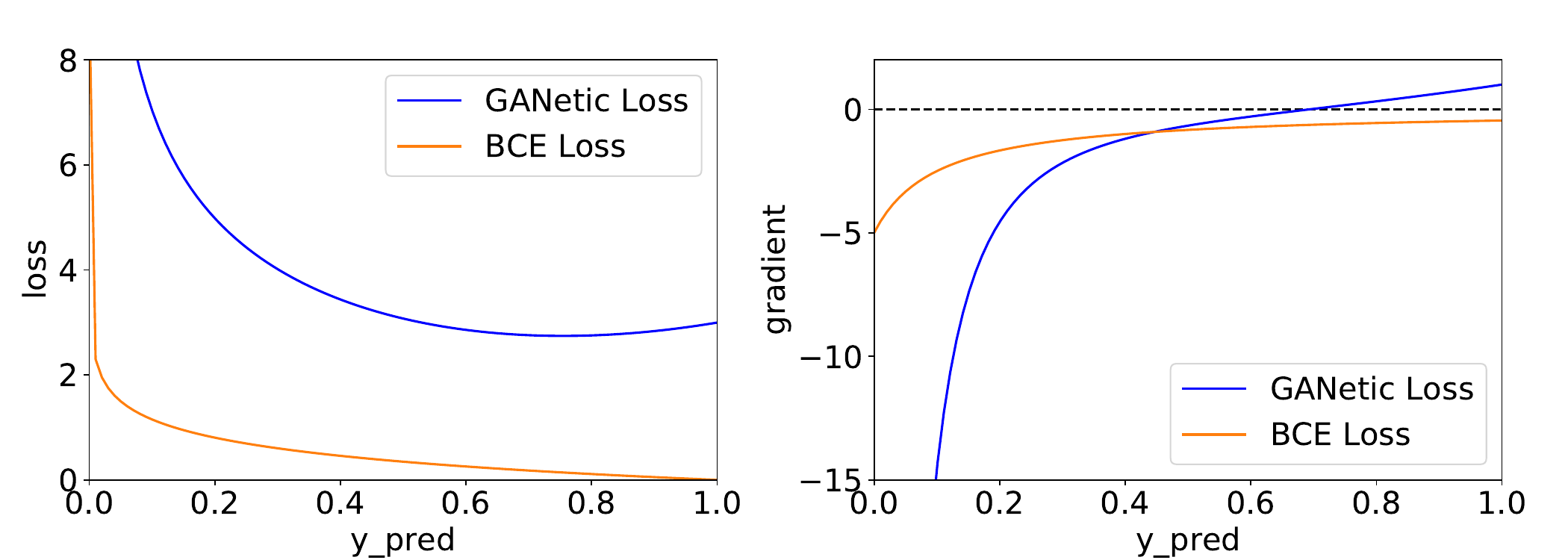}
\caption{GANetic and BCE loss functions (left) and their gradients (right).}
\label{fig:fig_8}
\end{figure}

The case where $y_{real}=1$ is plotted in Figure \ref{fig:fig_8} for the BCE and GANetic losses. The cross-entropy loss is shown to be monotonically decreasing, while the GANetic loss function shows an increase in the loss value as $y_{pred}$ approaches $y_{true}$. This increase of the loss value allows the GANetic loss function to prevent the model from becoming too confident in its output predictions and may provide an important advantage as it lowers the probability of overfitting. Thus, this loss function provides an implicit form of regularization, enabling better generalization.  

An implicit form of regularization naturally implemented in the GANetic loss as well as a prioritization of correctly classifying real images point directions of future work, as they can be used as prior knowledge introduced into GP or any other search paradigm to reduce the evaluation overhead. More comments can be found in Appendix \ref{app_discussion}.

\section{Conclusions}

In this study Genetic Programming was utilized for GAN loss function search. The final function was evaluated by using various GAN models and datasets. The obtained results of GANetic loss were compared to the ones achieved by the same models but trained by using other well-known loss functions such as binary cross entropy, hinge loss and Wasserstein loss. The GANetic loss was shown to convincingly outperform existing commonly used loss functions and to be efficient for network training regardless of model architecture or dataset. Moreover, it is applicable for medical problems such as image generation and anomaly detection.

However, while functions found by GP were achieved with limited populations size in this study, it is clear that experiments with GP with additional computing resources could lead to even better performing functions. In addition, separating generator and discriminator loss during the search process increases the search space and could lead to an even better, more specialized combined loss. Further analysis suggested that the improvements provided by the new loss result from implicit regularization that reduces overfitting to the data.

\section{Broader Impact}
Generative models learn the underlying representation of the training data, addressing a fundamental machine learning problem. Especially, in the medical sector those models pose a huge potential, for example allowing identification of novel disease phenotypes. In this study, we present an experimentally found loss function that generally improves the performance of GANs and additionally stabilizes the training procedure. Furthermore, we evaluated the loss for medical applications improving the performance of existing models. 

Despite the potential of generative models, they also pose a significant risk, particularly the generation of fake information. Moreover, there are risks in the medical field not only from misuse, but also from improper use, which can impair patient care. Nevertheless, we are convinced that the benefits outweigh the risks and may improve patient healthcare in the long run. Development of robust models and the detection or watermarking of fake images requires a community effort to further shift the balance towards the positive side. Our work does not provide fundamentally new capabilities, but rather improves the performance of existing models leading to more reproducible results using GAN-based networks. Our work may hopefully contribute in future work learning the representation of complex real-world phenomena.

\bibliography{example_paper}
\bibliographystyle{icml2024}

\newpage
\appendix

\section{Appendix: Genetic Programming}
\label{app_a}

The GP search process starts with random initialization of a set of potential solutions, which is also called a population of individuals, in the functional search space. The number of individuals is the predefined parameter $n$, which does not change through the whole search process. For this study a set of loss functions was randomly generated, and each loss function was an individual represented as a tree, where the terminals were randomly chosen from the set $\{y_{pred}, y_{real}, \mathbb{R}\}$, while the operators were chosen from the set $\{+, -, \times, \div, \sqrt, \log, \exp, \sin, \cos\}$. To utilize these functions and terminals for loss function generation, the following modifications were applied:
\begin{itemize}
\item $x \div y$ was changed to $x \div \left(y+\varepsilon\right)$, with $\varepsilon=10^{-8}$;
\item $\sqrt{x}$ was changed to $\sqrt{\left|x\right|+\varepsilon}$, with $\varepsilon=10^{-8}$;
\item $\log(x)$ was changed to $\log\left(\left|x\right|+\varepsilon\right)$, with $\varepsilon=10^{-8}$;
\item It was implemented as a rule that both $y_{pred}$ and $y_{real}$ should be present in the formula;
\item Regardless of the trained GAN's architecture, the sigmoid function was applied on the last layer so that $y_{pred}$ was always in the range $[0,1]$.
\end{itemize}

\section{Appendix: Parameter setting for GP}
\label{app_b}

\begin{table}[h!]
\renewcommand{\thetable}{B}
\caption{Parameter setting for GP.}
\label{tab:tab_1}
\centering
\begin{tabular}{lcccccc}
\toprule
 & $M_{ST}$ & $M_N$ & $p_A$ & $Cr_A$ & Selection & $k_t$ \\
\midrule
1	& $0.3$ & $0$ & $0$ & $0$ & tournament & $3$ \\
2	& $0.3$ & $0$ & $0.5$ & $0.5$ & tournament & $3$ \\
3	& $0.3$ & $0$ & $0$ & $0$ & select $n$ best & $-$ \\
4	& $0.3$ & $0$ & $0.5$ & $0.5$ & select $n$ best & $-$ \\   
5	& $0.2$ & $0.1$ & $0$ & $0$ & tournament & $3$ \\
6	& $0.2$ & $0.1$ & $0.5$ & $0.5$ & tournament & $3$ \\
7	& $0.2$ & $0.1$ & $0$ & $0$ & select $n$ best & $-$ \\       
8	& $0.2$ & $0.1$ & $0.5$ & $0.5$ & select $n$ best & $-$ \\
\bottomrule
\end{tabular}
\end{table}

The population size, $n$, is one of the most important hyperparameters. The population size, which was set at 10, determines the number of individuals in each generation, which evolve from generation to generation. The more individuals, the bigger is the pool of the potential solutions. However, GP is highly efficient even with the small populations, as the number of generations, $T$, which is another important hyper-parameter, highly affects the final results, as it provides more time for the initially randomly generated individuals to evolve to better fit the objective function. So although we set the population size to $10$, this was still sufficient for GP, as the other more important parameters were generally set very carefully. Additionaly, the computational cost scales with the number of individuals, so our choice was determined by the attempt to limit the computational workload of the conducted experiments.

\section{Appendix: Loss functions found for all GP experiments}
\label{app_c}

\begin{equation}
\tag{C1}
\label{eqn:eq2}
f_1=\frac{1}{N}\sum_{i=1}^{N}{\left(e^{2.2061}+\sin{(1.7577)}+(y_{real}^{(i)}-4.092)^3-\log{(|y_{real}^{(i)}-y_{pred}^{(i)}|+\varepsilon)}\right)}, \\
\end{equation}

\begin{equation}
\label{eqn:eq3}
\tag{C2}
f_2=\frac{1}{N}\sum_{i=1}^{N}{e^{\cos{\left(\sqrt{|y_{real}^{(i)}-y_{pred}^{(i)}|+\varepsilon}\right)}}},
\end{equation}

\begin{equation}
\label{eqn:eq4}
\tag{C3}
f_3=\frac{1}{N}\sum_{i=1}^{N}{\left(\sqrt{\left|y_{real}^{(i)}+\log{\left(\left|y_{pred}^{(i)}\right|+\varepsilon\right)}\right|+\varepsilon}\right)^3},
\end{equation}

\begin{equation}
\label{eqn:eq5}
\tag{C4}
f_4=\frac{1}{N}\sum_{i=1}^{N}{\left(\left(y_{pred}^{(i)}\right)^3+ \sqrt{\left|3.985 \times \frac{y_{real}^{(i)}}{y_{pred}^{(i)}+\varepsilon}\right|+\varepsilon}\right)},
\end{equation}

\begin{equation}
\label{eqn:eq6}
\tag{C5}
f_5=\frac{1}{N}\sum_{i=1}^{N}{\left[\left(\sqrt{\left|\log{\left(\left|y_{pred}^{(i)}\right|+\varepsilon\right)}\right|+\varepsilon}\right)^3+\frac{\log{\left(\left|\left(y_{pred}^{(i)}\right)^3\right|+\epsilon\right)}}{3.6278 \times y_{real}^{(i)} \times \left(y_{pred}^{(i)}\right)^2+\varepsilon}\right]},
\end{equation}

\begin{equation}
\label{eqn:eq7}
\tag{C6}
f_6=\frac{1}{N}\sum_{i=1}^{N}{\left(e^{\cos{\left(y_{pred}^{(i)}\right)}-\left(y_{pred}^{(i)}\right)^2} \times \left(y_{pred}^{(i)}-y_{real}^{(i)}\right)^4\right)},
\end{equation}

\begin{equation}
\label{eqn:eq8}
\tag{C7}
f_7=\frac{1}{N}\sum_{i=1}^{N}{\left[\sin{\left(1.0657 \times y_{real}^{(i)}+\frac{0.4129}{y_{pred}^{(i)}+\varepsilon}\right)}+\cos{\left(y_{real}^{(i)}+y_{pred}^{(i)}\right)^2}\right]},
\end{equation}

\begin{equation}
\label{eqn:eq9}
\tag{C8}
f_8=\frac{1}{N}\sum_{i=1}^{N}{\left[y_{pred}^{(i)} \times \cos{\left(y_{real}^{(i)}\right)}^2 \times \log{\left(\left|y_{real}^{(i)} \times y_{pred}^{(i)}\right|+\varepsilon\right)}-\sqrt{\left(\left|\log{\left(\left|y_{pred}^{(i)}\right|+\varepsilon\right)}\right|+\varepsilon\right)}\right]}.
\end{equation}

\section{Appendix: Discriminator Accuracy for GAN Performance Evaluation}
\label{app_mnist}

The discriminator values should only be considered in conjunction with the FID values of the generated images. First of all, the discriminator should have a score between 50\% and 100\%, and would otherwise indicate a corrupt discriminator. Secondly, given the same FID values a higher discriminator accuracy points to a better generative performance, as having a discriminator that is still able to discriminate fake and generated samples has the potential to further improve the generator. A discriminator being unable to discriminate generated from real samples cannot point the direction for the generator to produce better results and thus may not further improve generative results. 

The CIFAR10 results (Section \ref{cifar10}) showed a discriminator accuracy of about 50\% for GANetic loss indicating that training finished properly using the full discriminator potential. As a side note, we did train GANetic loss on different classification tasks (CIFAR10, ImageNet) where it showed similar performance in classification accuracy compared to cross entropy loss.

\section{Appendix: Evaluation of the GANetic Loss on CIFAR10 Dataset}
\label{app_cifar10}

\begin{figure}[h!]
\renewcommand{\thefigure}{E}
\centering
\includegraphics[width=1.0\linewidth]{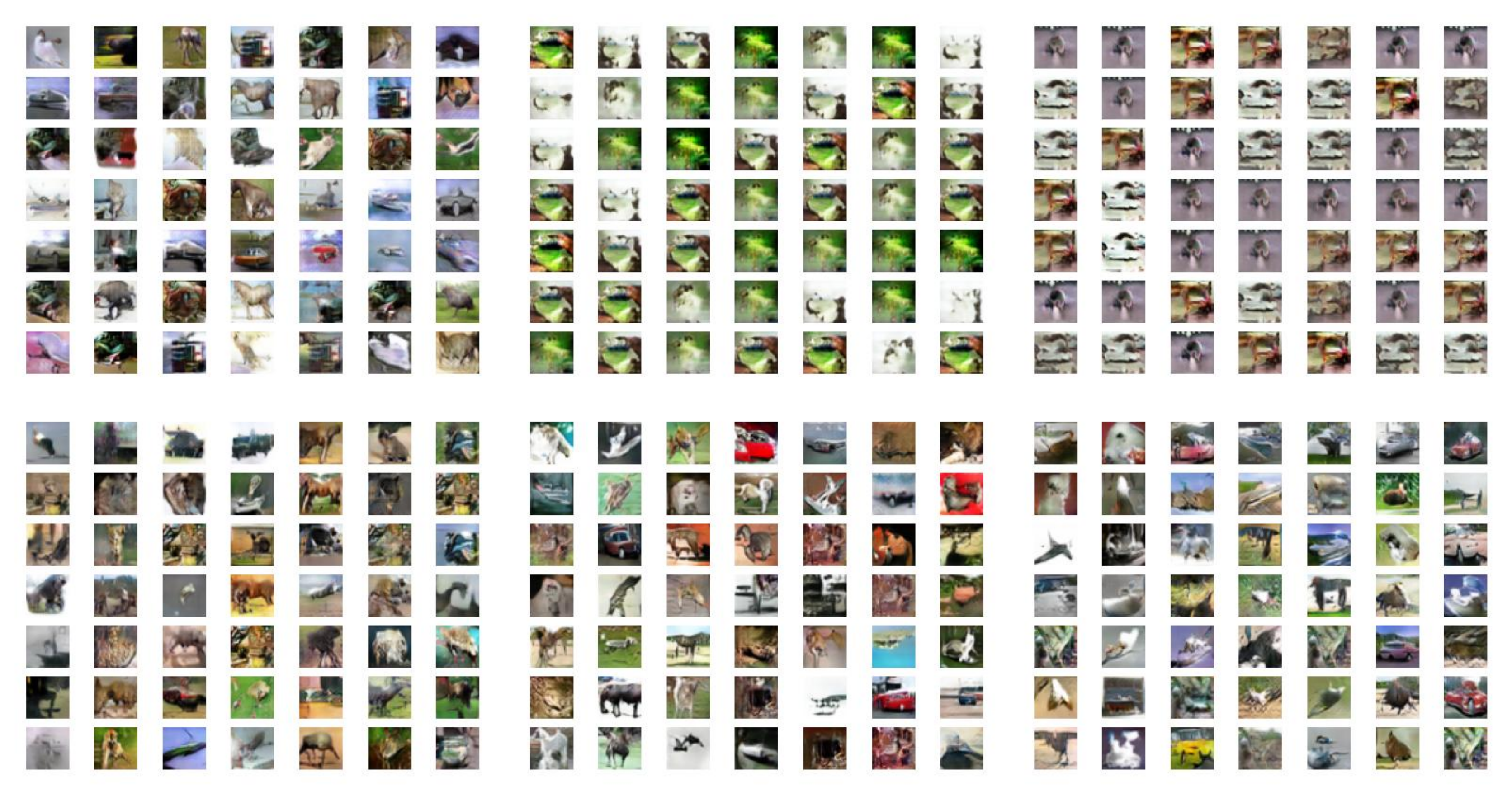}
\caption{Examples of images generated by DCGANs for CIFAR10 dataset. GANs trained by BCE (top) or trained by GANetic loss (bottom).}
\label{fig:fig_cifar10}
\end{figure}

In Figure \ref{fig:fig_cifar10}, images generated with GANetic loss and BCE (as it was the second best on CIFAR10 dataset) are compared.

\begin{table}[h!]
\renewcommand{\thetable}{E}
\caption{Results of ablation study on CIFAR10 for GANetic loss.}
\label{tab:tab_5}
\centering
\begin{tabular}{lccc}
\toprule
 Generator & Discriminator &  Mean FID & STD\\
\midrule
\checkmark && 9.67&3.60\\
&\checkmark&2.51&0.57\\
\checkmark &\checkmark& \textbf{2.20}&\textbf{0.39}\\
\bottomrule
\end{tabular}
\end{table}

To be more specific, in Figure \ref{fig:fig_cifar10} the first column of images in the series was generated during the best run of the DCGAN model used for BCE and GANetic loss, the second column of images was generated from the worst DCGAN, and the third column of images was generated from the model randomly selected from the remaining program runs. DCGANs trained with the BCE loss function produced the best results compared to other regular losses, but were still unable to produce images indistinguishable to those of the original dataset, while DCGANs trained with GANetic loss produced convincing images in all experiments.

\section{Appendix: Convergence of Projected GANs}
\label{app_d}

\begin{figure}[h!]
\renewcommand{\thefigure}{F}
\centering
\includegraphics[width=0.75\linewidth]{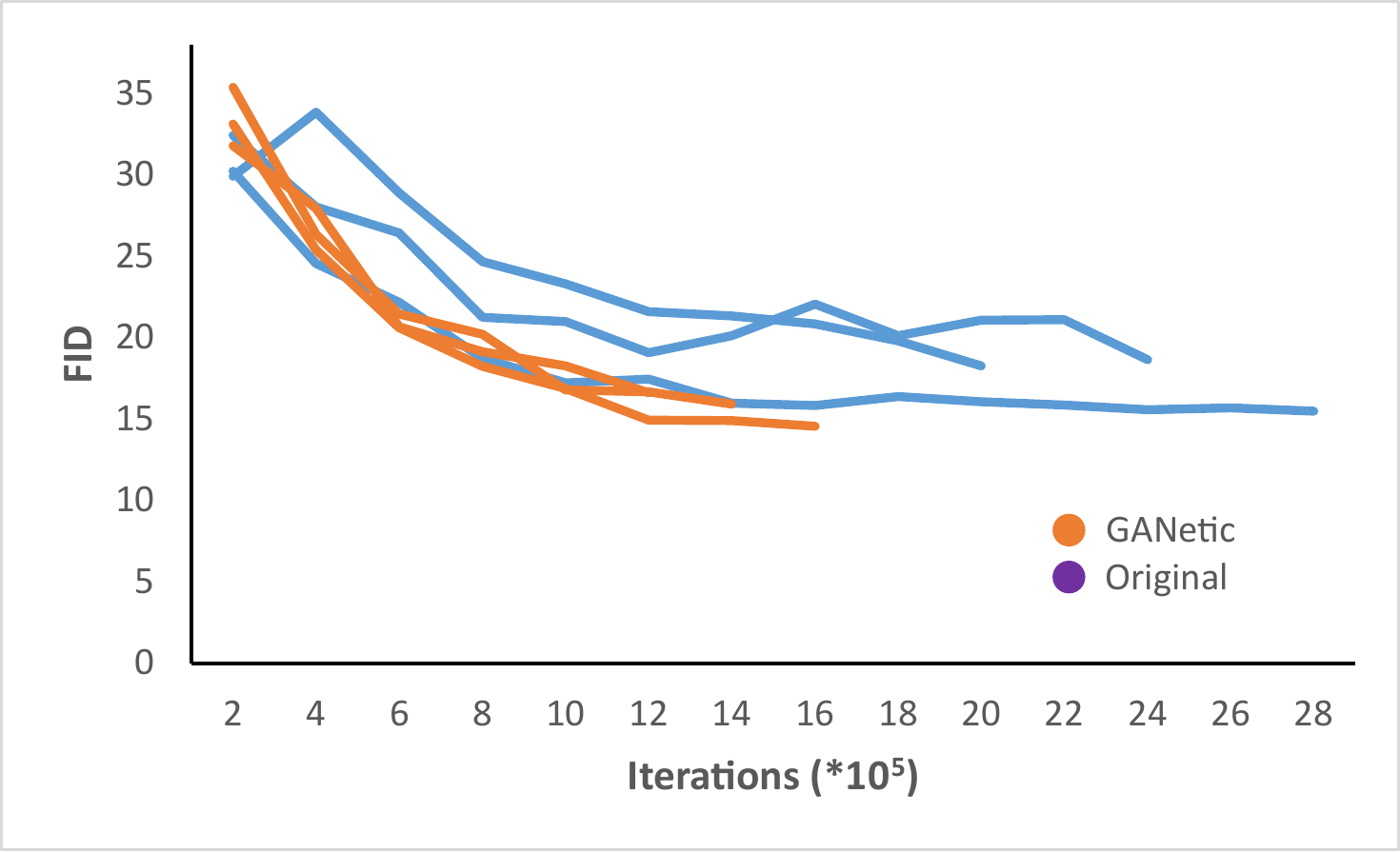}
\caption{Convergence of prGANs trained by GANetic, BCE and Hinge losses on the BreCaHAD dataset.}
\label{fig:fig_conv}
\end{figure}

\newpage

\section{Appendix: Image Generation for Medical Applications}
\label{app_med_image_gen}

\begin{figure}[h!]
\renewcommand{\thefigure}{G}
\centering
\includegraphics[width=\columnwidth]{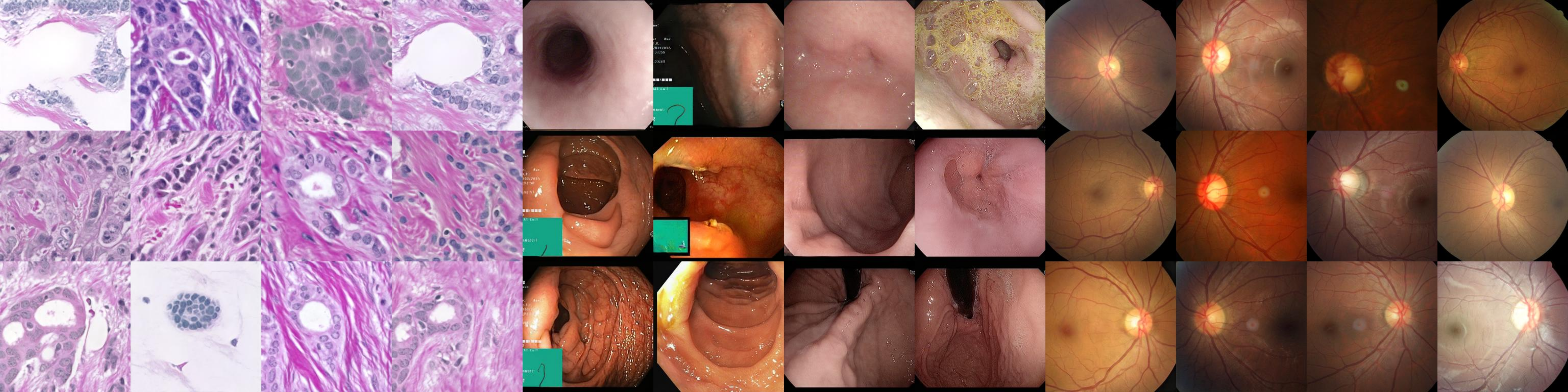}
\caption{Examples of images generated by StyleGAN2-ADA using GANetic loss for BreCaHAD (left), Hyper-Kvasir (middle) and LAG (right) datasets.}
\label{fig:fig_style}
\end{figure}

\section{Appendix: Additional Evaluation of the GANetic Loss.}
\label{app_afhq}

\begin{table}[h!]
\caption{FID and KID values for different losses trained on AFHQ-Cats,AFHQ-Dogs and AFHQ-Wild datasets ~\cite{choi2020stargan}.}
\label{tab:tab_afhq}
\renewcommand{\thetable}{H}
\centering
\begin{tabular}{lcccccc}
\toprule
Model (Loss)  & \multicolumn{2}{c}{AFHQ-Cats} & \multicolumn{2}{c}{AFHQ-Dogs} & \multicolumn{2}{c}{AFHQ-Wild} \\
\cmidrule(r){2-7}
& FID & KID & FID & KID & FID & KID \\
\midrule
prGAN (original) & 3.25 & 0.29 & 7.08 & 2.12 & 4.21 & \textbf{0.20} \\
prGAN (GANetic) & \textbf{2.64} & \textbf{0.19} & \textbf{6.69} & \textbf{1.86} & 4.86 & 0.35 \\
StyleGAN2-ADA (original) & 3.55 & 0.66 & 7.40 & 1.16 & 3.05 & 0.45 \\
StyleGAN2-ADA (GANetic) & 3.22 & 0.61 & 9.12 & 2.33 & \textbf{2.38} & 0.31 \\
\bottomrule
\end{tabular}
\end{table}

\section{Appendix: Anomaly Detection}
\label{app_e}

\begin{table}[h!]
\renewcommand{\thetable}{I}
\caption{Ablation study of f-AnoGAN configuration (D: Discriminator, G: Generator, E: Encoder), values are mean AUROC $\pm$ STD.}
\label{tab:tab_9}
\centering
\begin{tabular}{lcccc}
\toprule
$D$  & $G$ & $E$ &Kvasir & LAG \\
\midrule
\checkmark &  &  & \textbf{$99.01 \pm 0.06$} & $78.03 \pm 1.19$ \\
 & \checkmark &  & $93.10 \pm 5.24$ & $75.39 \pm 1.32$ \\
 &  & \checkmark & $76.06 \pm 0.69$ & $58.39 \pm 1.62$ \\
\checkmark & \checkmark &  & $98.97 \pm 0.14$ & \textbf{$78.22 \pm 0.88$} \\
 &\checkmark & \checkmark & $76.58 \pm 0.28$ & $59.29 \pm 0.80$ \\
\checkmark &  & \checkmark & $76.92 \pm 0.11$ & $57.75 \pm 0.47$ \\
\checkmark & \checkmark & \checkmark & $76.99 \pm 0.03$ & $58.50 \pm 1.91$ \\
\bottomrule
\end{tabular}
\end{table}

\begin{figure}[h!]
\renewcommand{\thefigure}{I}
\centering
\includegraphics[width=1.0\linewidth]{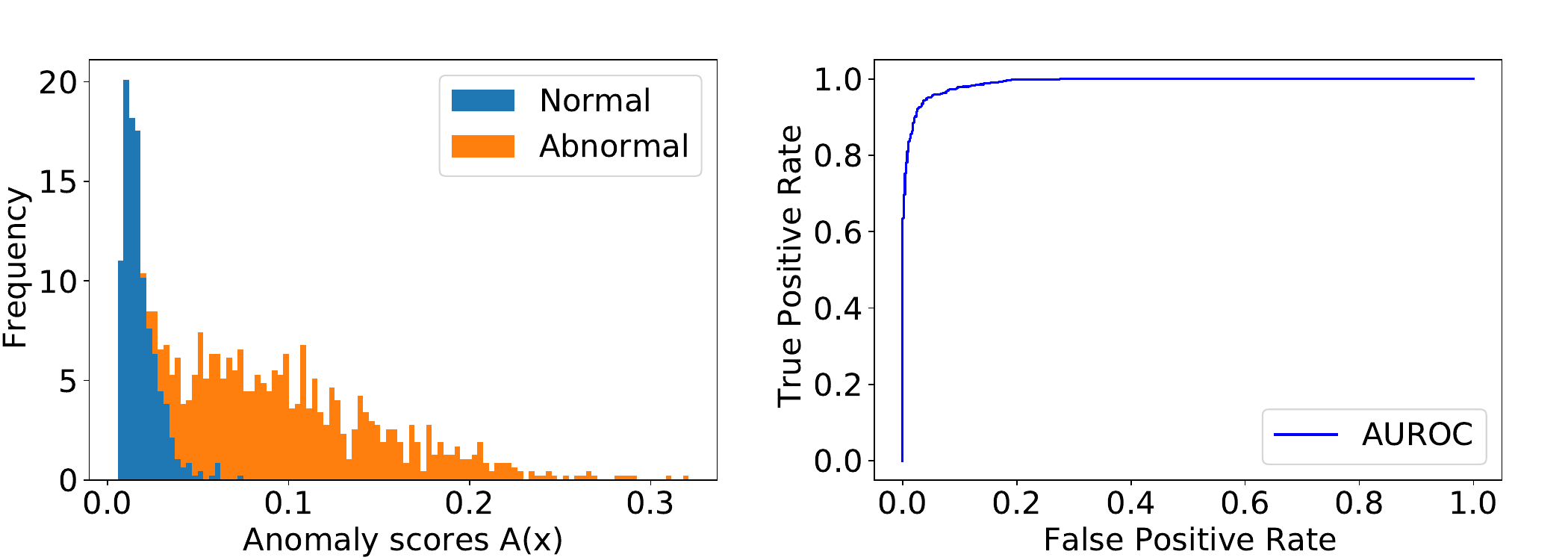}
\caption{Discrete distributions of anomaly scores for the Hyper-Kvasir dataset obtained by the f-AnoGAN model, trained using GANetic loss.}
\label{fig:fig_kvasir}
\end{figure}

\section{Appendix: Discussion}
\label{app_discussion}

Generally, the cross entropy loss has a similar shape as GANetic loss, but BCE is monotonically decreasing. Additionally, the $f_6$ loss (Equation \ref{eqn:eq7}), that is the second best in Table \ref{tab:tab_2}, has a very similar shape compared to the GANetic loss when $y_{pred}=1$: it also has a minimum and after reaching it the loss value increases even though the model's confidence in the prediction is high. The minimum of the $f_6$ loss is at $1.01$, making it monotonically decreasing in the range $[0,1]$. Thus, its behaviour is the same as the one of the BCE loss in the range $[0,1]$. However, it wasn't able to train DCGAN on CIFAR10 dataset, which led to its exclusion from the further experiments. 

To evaluate the implicit regularization more comprehensively a more systematic approach would be required with an in-depth analysis that would be out of scope of the current work. However, the slope of the GANetic function could be the starting point for such an analysis in future work and may lead to further theoretical insights about implicit regularization and optimization of adversarial networks.




\end{document}